\documentclass{article}
\usepackage{ijcai23}

\usepackage{times}
\usepackage{soul}
\usepackage{url}
\usepackage[hidelinks]{hyperref}
\usepackage[utf8]{inputenc}
\usepackage[small]{caption}
\usepackage{graphicx}
\usepackage{amsmath}
\usepackage{amsthm}
\usepackage{booktabs}
\usepackage{algorithm}
\usepackage{algorithmic}
\usepackage[switch]{lineno}

\linenumbers

\title{Balancing Accuracy and Training Time in Federated Learning for Violence Detection in Surveillance Videos: A Study of Neural Network Architectures}

\author{
PAJON Quentin$^1$
\and
SERRE Swan$^1$\and
WISSOCQ Hugo$^1$\and
RABAUD Léo$^1$\and
HAIDAR Siba$^{1,2}$\And
YAACOUB Antoun$^{1,2}$
\affiliations
$^1$ESIEA, Paris, France\\
$^2$Learning, Data and Robotics (LDR) Lab, ESIEA, Paris, France
\emails
\{pajon, sserre, wissocq, lrabaud\}@et.esiea.fr, \{siba.haidar, antoun.yaacoub\}@esiea.fr
}

\begin{document}
\nolinenumbers
\maketitle

\begin{abstract}
  This paper presents an investigation into machine learning techniques for violence detection in videos and their adaptation to a federated learning context. The study includes experiments with spatio-temporal features extracted from benchmark video datasets, comparison of different methods, and proposal of a modified version of the "Flow-Gated" architecture called "Diff-Gated." Additionally, various machine learning techniques, including super-convergence and transfer learning, are explored, and a method for adapting centralized datasets to a federated learning context is developed. The research achieves better accuracy results compared to state-of-the-art models by training the best violence detection model in a federated learning context.
\end{abstract}

\section{Introduction}
Violence detection can be used in many contexts: soccer stadiums, surveillance cameras, video sharing services, etc.
Moreover, humans aren't able to detect violence on this scale because of the huge quantity of data involved.
In the context of a CCTV center such AI can be used to inform the authorities and permit to intervene faster  \cite{youssef2021automatic}.

Federated learning can play a crucial role in ensuring data privacy and security for violence detection in surveillance videos, especially in compliance with GDPR regulations \cite{GDPR}. By keeping the training data and machine learning model on the devices where it was collected, the risk of sensitive information being intercepted on the network can be avoided \cite{gosselin_privacy_2022}. Moreover, federated learning can leverage techniques like differential privacy \cite{diffPrivacy} to protect individual privacy and reduce the risk of model reverse-engineering \cite{cheng2020federated}. This approach can be particularly useful in contexts such as CCTV centers, where timely detection of violence can inform authorities and help to intervene faster. 

The ethical implications tied to the use of machine learning for violence detection in surveillance contexts are significant. Potential issues range from infringements of privacy rights \cite{Crawford2013BigDA} to biases in the algorithm, potentially leading to discriminatory outcomes \cite{zou2018ai}, \cite{buolamwini2018gender}. These systems, while technologically innovative, need to be used judiciously to avoid misuse and ensure fairness and transparency \cite{mittelstadt_ethics_2016}, \cite{greene2019better}. They should not replace human judgement, but act as supportive tools for security personnel. Thus, it becomes critical to foster responsible AI practices, placing paramount importance on individual privacy and fairness.

In this paper, we propose a deep learning architecture for violence detection that can be effectively trained using federated learning, with memory efficiency and reduced training time as key considerations. By exploring the use of spatio-temporal features and machine learning techniques such as super-convergence and transfer learning, we achieve better accuracy results compared to state-of-the-art models. Our approach can pave the way for privacy-preserving and efficient violence detection in a range of real-world scenarios.

\section{Related work}
In this section, we explore two important topics in the field of data science: violence detection and federated learning. 

\subsection{Violence Detection}
Violence detection is a sub-field of action recognition that consists of detecting specific actions in videos. The detection of violence or fight scenes has been an active research field for a long time. Classical methods for violence detection used hand-crafted features, such as ViF (Violent Flow) \cite{hassner_violent_2012}, which detects changes in optical flow, and OViF (Oriented Violent Flow) \cite{gao_violence_2016}, an improvement of ViF that makes better use of the orientation information of optical flows. Optical flow is the apparent changes in the pixels of two consecutive images of a video.

More recently, deep learning methods have been developed, achieving better results and requiring less processing than classical methods, allowing the model to learn the violence patterns. To do this, several architectures have been experimented with over time. One method is to use 3D CNNs \cite{ding_violence_2014}, which applies convolutions to videos. This method has improved over time, such as in \cite{li_efficient_2019}, which applies the concepts of DenseNet \cite{huang_densely_2018} to improve the model's performance and reduce the number of parameters.

Another deep learning method is to use Conv-LSTM cells \cite{shi_convolutional_2015} and a traditional CNN. LSTM aggregates the features extracted by the CNN on the frames of the video to obtain temporal information. This is what is done by \cite{sudhakaran_learning_2017}.

\cite{cheng_rwf-2000_2020} proposed using two input channels instead of one corresponding to the video frames. The additional channel is the optical flow, which helps the model focus on the areas where there is movement and possibly violence.

\subsection{Federated Learning}
Data governance has become an important consideration in violence detection research, particularly with regards to the ethical and legal implications of video data privacy following the implementation of the GDPR \cite{GDPR}. In response, federated learning has emerged as a promising approach for training data science models without centralized data storage, with several algorithms such as FederatedAveraging \cite{mcmahan_communication-efficient_2017}, FedProx \cite{li_federated_2020}, MIME \cite{karimireddy_breaking_2021}, and FEDOPT \cite{reddi_adaptive_2021} proposed for this purpose.

In this paper, we adopt FederatedAveraging \cite{mcmahan_communication-efficient_2017}, a technique that allows each client to train a local model with their own data. The resulting models are then averaged to train a global model that is sent back to the clients after each round. FederatedAveraging has been demonstrated to be well-suited to Non-IID (independent and identically distributed) data distributions in \cite{mcmahan_communication-efficient_2017}.

Relatedly, \cite{victor2022federated} work in 2022 also leveraged federated learning for violence detection in videos, using various pre-trained convolutional neural networks on the AIRTLab Dataset and finding the best performance with MobileNet architecture, showcasing a high accuracy of 99.4\% with only a marginal loss when compared to non-federated learning settings.

% \subsection{Challenges and Limitations of Federated Learning for Violence Detection}

% Although our work has demonstrated the feasibility of federated learning for violence detection, there are inherent challenges and limitations to this approach. One of the primary issues is the uneven distribution of data across devices, known as non-IID data, which can lead to significant performance degradation \cite{li2020federated}. Moreover, the computational and storage capabilities of client devices can be vastly different, leading to potential bottlenecks in the training process \cite{kairouz2019advances}.

% Another challenge is the potential increase in communication overhead, as federated learning requires exchanging model updates between the server and clients. This can lead to higher latency and energy consumption, particularly in scenarios with limited bandwidth or large model sizes \cite{bonawitz2019towards}. 

% Lastly, privacy, although improved in federated learning, remains a challenge. Malicious parties could potentially infer sensitive information from the shared model updates, necessitating careful design of privacy-preserving mechanisms \cite{geyer2017differentially}.

\section{Methodology}

 In this subsection, we present the architectures of the models used for violence detection and how we adapted these models for the federated learning context. Our objective is to achieve state-of-the-art or higher accuracy while being resource and time-efficient since these models are intended to be deployed and trained in CCTV centers, which typically have less powerful computers than research environments. We also describe the methodology we used to develop and optimize our violence detection model, which involved addressing the limitations of classical classifiers using transfer learning, early stopping, and One-Cycle training. Additionally, we explore multi-channel input models using both optical flow and frame differences, and detail the specific model architectures we used to achieve better performance while reducing computation time. Overall, this section provides a comprehensive overview of our approach to developing a robust and efficient violence detection model.

\subsection {Limitation of Classical Classifiers}

\begin{figure*}[!ht]
\centering
\includegraphics[width=1\textwidth]{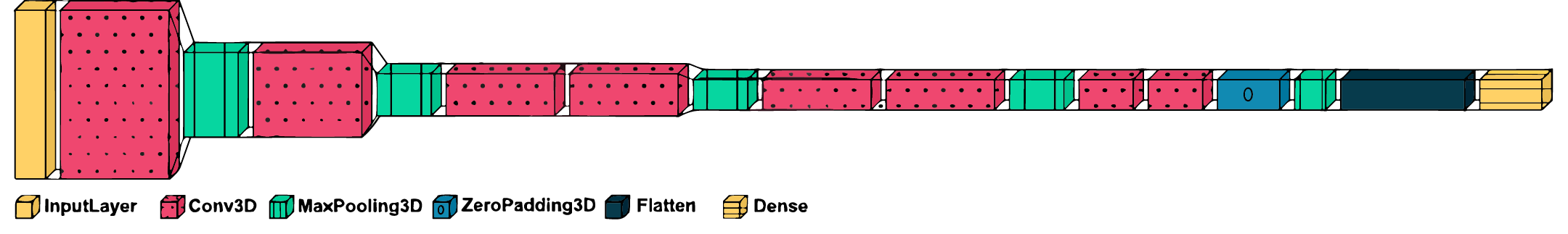}
\caption{\label{fig:Feature_extraction_model}Structure of the feature extraction architecture \protect\cite{sernani_deep_2021}}
\end{figure*}

\cite{sernani_deep_2021} utilized a pre-trained model on the Sports-1M dataset \cite{tran_learning_2015} for feature extraction (see Figure  \ref{fig:Feature_extraction_model}) and employed an SVM classifier to classify videos as violent or non-violent.

We explored different classification methods for violence detection and evaluated their performance. In particular, we tested decision tree and random forest classifiers using a pre-trained model to extract video features. We optimized the hyperparameters of both classifiers using grid search and trained them on the extracted features. Despite being marginally faster  than the SVM classifier, the decision tree and random forest classifiers did not meet our needs in terms of speed and accuracy on our dataset. For instance, it took around 5 minutes to extract features from 300 videos of 5 seconds in 30 fps and low resolution, followed by 10 seconds for the classifier training, which is not efficient for our use case.

This experimentation revealed that classical classification methods on extracted features for violence detection are not well-suited for real-time applications.

\subsection {Transfer Learning and Early Stopping}\label{sec3.2}

To improve training efficiency, it was essential to explore various transfer learning architectures. 

\begin{figure*}[htbp]
\centering
\resizebox{2\columnwidth}{!}{\includegraphics{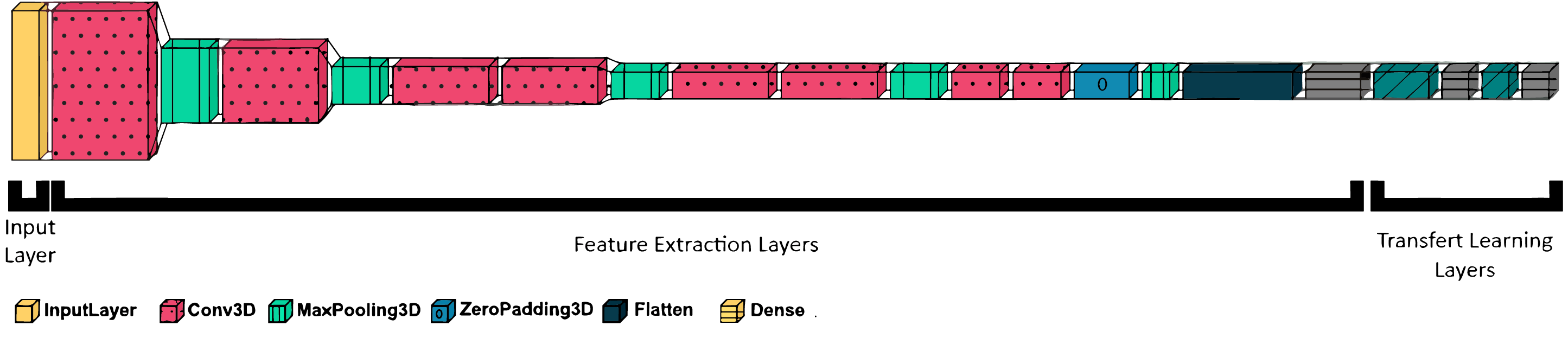}}
\caption{\label{fig:transfer_learning} Structure of the transfer learning architecture \protect\cite{sernani_deep_2021}}
\end{figure*}

\cite{sernani_deep_2021} utilized a transfer learning architecture consisting of six pre-trained layers, with two fully connected layers of 4096 and 512 neurons respectively, and a final output layer (see Figure \ref{fig:transfer_learning}).

We tried many different architectures, from starting the transfer learning one layer earlier, to adding multiple training layers, each layer containing from 256 to 4096 neurons. The goal being to get a better training time or ROC curve than the model used in the previous paper, therefore, the time to beat was 12min30s and the AUC was 0,987.
Through trial and error, we obtained results ranging from 8 minutes of training time and an AUC of 0.941 to 17 minutes and and AUC of 0.991. 
We noticed that the more neurons we used on the before last layer, the faster the network converged, but at the same time, the worse the accuracy became.
However, these results are not what we need since they are slower and less accurate than what a random forest classifier is able to produce (Table \ref{tab:classifiers}). The aim being to get the most efficient training possible due to material constraints in a federated context, we want to reduce training time as much as possible, which leads us to the next section.

\subsection{\label{Model qui federe}Transfer Learning and One-Cycle Training}

To reduce training time without sacrificing results, we aimed to accelerate the convergence of our networks. We used a technique called cyclical learning rates, introduced by \cite{smith_cyclical_2017}, which involves gradually increasing the learning rate from a very low value to a high value and then decreasing it back to the starting value. This technique can help speed up convergence.

The super-convergence \cite{smith_super-convergence_2018} uses a single cycle of cyclical learning with optimal learning rates to train a model for a few epochs. This method is called One-Cycle training. In order to find these lower and higher bounds of learning rates, we trained the model for one single epoch with a learning rate gradually rising from $1e^{-15}$ to 1 and plotted the loss on a graph as the Y-axis with the learning rate as the X-axis. The minimum and maximum learning rates were respectively located after a sharp drop and before a sharp rise in loss. We used this method to reduce training time while maintaining high accuracy and ROC curves with an Area Under Curve of over 0.96.

We, then, experimented with various architectures by varying the number of layers, number of neurons, and starting layer for transfer learning. We found that training time evened out completely, and that lighter architectures starting from the seventh layer tended to produce better results. Therefore, we chose an architecture with 1024 neurons on the seventh layer and a final fully connected layer (see Figure \ref{fig:transfer_learning}). This approach allows for faster training times than feature extraction methods while maintaining similar accuracy.

\subsection{Multi-Channel Input Models using Optical Flow}

\begin{figure*}[htbp]
\centering
\resizebox{2\columnwidth}{!}{\includegraphics{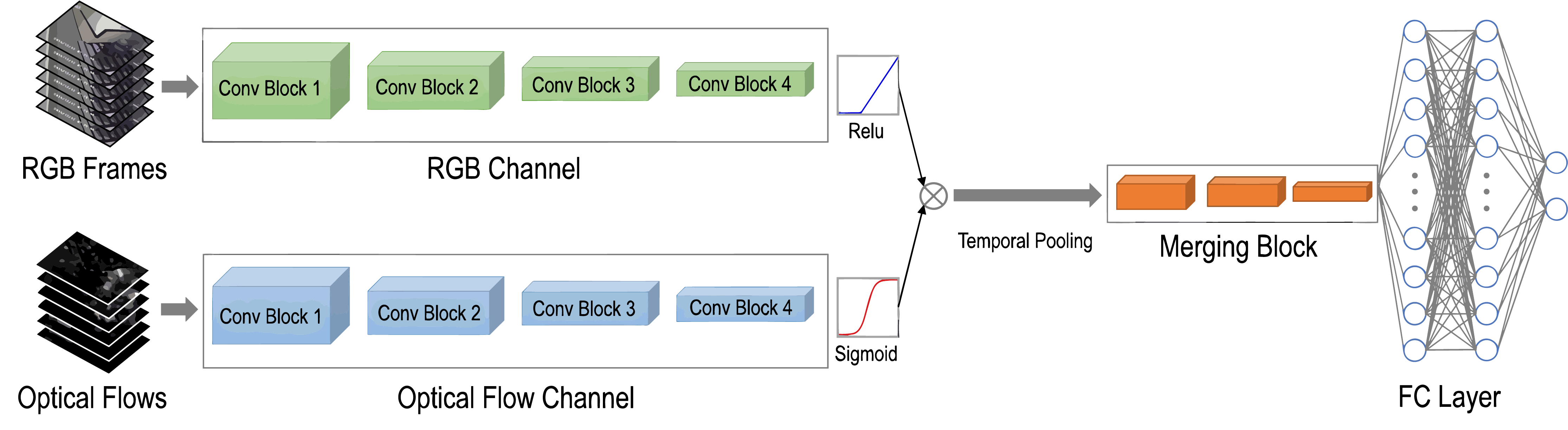}}
\caption{\label{fig:flow-gated} Structure of the flow gated architecture \protect\cite{cheng_rwf-2000_2020}}
\end{figure*}

We experimented with multi-input models, including the Flow-Gated architecture proposed by \cite{cheng_rwf-2000_2020}, which has a lightweight design (272,690 parameters) and uses both RGB frames and optical flow inputs to identify areas of movement and potential violence (see Figure \ref{fig:flow-gated}). The model consists of four blocks, including the RGB and Optical Flow Channels, a Merging Block, and a Fully Connected Layer, with all 3D CNN blocks using depth-wise separable convolutions from MobileNet \cite{howard2017mobilenets} and Pseudo-3D Residual Networks \cite{qiu2017learning} to reduce parameters without sacrificing performance. However, the time required to compute optical flow is a significant drawback, with an average computation time of 9 seconds for a 5-second, 30fps video, making it unsuitable for near real-time violence detection and minimizing learning time.

\subsection{Multi-Channel Input Models using Frame Differences}

To reduce computation time required for calculating optical flow, we opted to use frame differences instead, as demonstrated in \cite{islam_efficient_2021}. This method requires significantly less computation time (0.065 seconds compared to 9 seconds for optical flow) and also reduces the number of input channels required. Unlike optical flow, which also provides direction of movement, frame differences only detect changes in the image.

We retained the original Flow-Gated model architecture and modified the optical flow channel to use frame differences. We also added more dropout to the fully connected layer to prevent overfitting. This modified architecture, which we named "Diff-Gated", is lightweight (272,546 parameters) and achieves better results than the original Flow-Gated network, as shown in Table \ref{table:gated_results}. These changes have reduced preprocessing and training time while improving accuracy. 

\section{Experiment}
In this section, we describe the experiments we conducted to develop and optimize our violence detection model. We begin by explaining our choice of dataset and the setup we used for our experiments. We then discuss how we adapted the dataset to the federated learning context and the challenges we faced. In the next section, we present our results for training previously tested models in a federated setting.
\subsection{Dataset Selection}

Various datasets exist for violence detection, such as the Movies dataset \cite{real_violence_2011} consisting of fight scenes from movies or the Hockey Fight dataset \cite{real_violence_2011} composed of fights from hockey games. However, these datasets have specific video contexts and may not represent real-life situations. To choose the most relevant dataset, we looked for CCTV footage of real-life physical violence situations with a significant number of videos and that have been used in other studies.

We selected the RWF-2000 dataset \cite{cheng_rwf-2000_2020}, consisting of 2000 videos from various sources of CCTV cameras, with 1000 violent videos and 1000 non-violent videos. All videos are 5 seconds long and filmed at 30 fps. To ensure the reliability of our models in multiple situations, we also used three other datasets: the crowd violence dataset \cite{hassner_violent_2012}, the AIRTLab dataset \cite{bianculli_dataset_2020}, and the hockey fights dataset \cite{real_violence_2011}.

However, training the RWF-2000 dataset required a significant amount of RAM, even with high-memory resources like Google Colab (83.5GB RAM). To avoid memory issues, we limited ourselves to using 300 to 400 videos for training.

\subsection{Experimental Setup: Hardware and Software}

All our experiments were conducted using Google Colab. For the non-federated models, we used a high-RAM setup and a standard GPU (Tesla T4). For the federated model experiments, we used a high-RAM setup with a premium GPU (Nvidia V100 or A100), as the power and RAM associated with a premium GPU were necessary for federated learning. We needed to simulate multiple clients and a server on a single machine to create a federated learning environment.

Given the novelty of our engagement with Google Colab and video classification, coupled with the challenging timeline, our primary focus was directed towards other pivotal aspects of the project. Consequently, the optimization of memory usage presents an area for further exploration and improvement.

\subsection{\label{adaptation_lol}Data Preparation for Federated Learning: Adapting Traditional Datasets}

The datasets we have chosen for violence detection consist of videos labelled either "violent" or "non-violent". However, in a federated learning context, multiple "clients" are needed, each simulating a different data source. This presents a challenge, as traditional datasets are not designed for federated learning.

To address this challenge, we propose and develop a method to simulate a federated-learning ready dataset from a traditional one. Our method involves a stratified split, which means that each video is associated with one and only one client, and each client has approximately the same number of violent and non-violent videos. This ensures that the data distribution is balanced across the different clients and reduces the risk of Non-IID data, making it easier to train a model.

Our proposed method offers several advantages. Firstly, it enables us to use traditional datasets for federated learning. Secondly, it helps to prevent the failure of models due to Non-IID data distribution. Lastly, it provides a more realistic representation of the data distribution in real-world scenarios, where different data sources may have varying proportions of violent and non-violent videos. 

\subsection{Federated Learning with Previously Tested Models}

We focused on training only one of our models in a federated context. Our choice was the model presented in \ref{Model qui federe} because it had the best accuracy/training time ratio while reducing the number of parameters.

We experimented with two frameworks: TensorFlow Federated \cite{noauthor_tensorflow_federated_nodate} and Flower \cite{beutel_flower_2022}. While achieving an accuracy of over 80\% with TensorFlow Federated is extremely time-consuming, we found it easier to do so with Flower. As a result, we decided to focus our efforts on Flower.

Although we were satisfied with our metrics using Flower, we encountered issues because no memory was freed between training rounds. Despite this, we were able to successfully train the model presented in \ref{Model qui federe} using the FederatedAveraging algorithm \cite{mcmahan_communication-efficient_2017}.

We observed the following:
\begin{itemize}
\item When training on a random sample of data sources each round, our metrics were slightly lower than those obtained outside of a federated context. We consider this result expected because the model was trained on fewer data per round.
\item When training on every data source each round, the accuracy was higher than that obtained outside of a federated context. We believe this is due to the multiple rounds of training, corresponding to multiple complete training cycles, instead of two epochs used for training using the One-Cycle method.

\end{itemize}

In conclusion, we were able to successfully adapt our non-federated models to a federated learning context using the Flower framework. We were able to train the model presented in Section \ref{Model qui federe} using the FederatedAveraging algorithm. Although we faced some challenges due to memory size issues in a federated context, we were still able to achieve good metrics for our federated model. Our results show that training on a random sample of data sources each round resulted in slightly lower accuracy, while training on every data source each round resulted in higher accuracy compared to training outside of a federated context. These findings suggest that federated learning has the potential to improve violence detection models while preserving privacy, and could be further explored in future work.

\section{Results}

In this section, we present the results of our experiments on violence detection in videos using deep learning techniques. We first compare the accuracies of different classifiers following feature extraction on three hundred videos of the RWF-2000 database. We also report the preprocessing times of different datasets, which are influenced by video framerate, resolution, and length. Next, we provide the results of our proposed model trained on four different datasets, with different numbers of epochs and maximum learning rates. We also compare the accuracy and computation time of our Diff-Gated model with the original Flow-Gated architecture on the RWF-2000 dataset. Finally, we show the validation accuracy of our federated learning model on the RWF-2000 dataset for each round of training.

\subsection{Classifiers Comparison}
Table \ref{tab:classifiers} presents the accuracy and training times of various classifiers using the C3D model and different feature extraction methods on 300 videos of the RWF-2000 dataset \cite{cheng_rwf-2000_2020}. The feature extraction process takes approximately five minutes, which needs to be added to the training time to determine the overall time required for generating a classifier.

\begin{table}
    \centering
    \begin{tabular}{lrr}
        \toprule
        Classifier & Accuracy & Training time  \\
        \midrule
        C3D + SVM(fc7) & 99.8\%   & 23s \\
        C3D + Decision Trees(fc7) &  83.9\%  &  10s   \\
        C3D + Random Forest (fc7) & 99.5\% &  10s  \\
        C3D + Decision Trees(fc6) &   85.9\% &  10s  \\
        C3D + Random Forest(fc6) &  99.5\% & 10s  \\
        \bottomrule
    \end{tabular}
    \caption{Accuracy and training times of the extraction feature methods on 300 videos of the RWF-2000 dataset.}
    \label{tab:classifiers}
\end{table}

\subsection{Establishing Baseline Model Performance for Subsequent Comparisons}

Although using decision trees instead of a SVM can save some time in training, the amount of time needed to extract features makes it not worthwhile to pursue this avenue in our context \cite{sernani_deep_2021}. The preprocessing times shown in table \ref{tab:datasets} are influenced by several factors, such as the frame rate, resolution, and length of the videos used in the datasets. For instance, the Hockey Fights dataset takes less time to process because each video lasts only one second, is filmed at around 30fps, and has a resolution of 360p. On the other hand, the AIRTlab dataset has longer videos, with a duration of five seconds, filmed at 30fps and a resolution of 1080p.

\begin{table}
    \centering
    \setlength\tabcolsep{4pt}
    \begin{tabular}{lrr}
        \toprule
        Dataset & \# videos & $\approx$ preprocessing time \\
        \midrule
        AIRTLab  & 350 & 3m30s \\
        Crowd Violence  & 246 & 36s \\
        Hockey Fights  & 1000 & 1m08s\\
        RWF-2000 (400 videos)  & 400 & 2m36s\\
        \bottomrule
    \end{tabular}
    \caption{Number of videos per dataset and their approximate preprocessing time.}
    \label{tab:datasets}
\end{table}

The results presented in Table \ref{tab:datasets_results} can be used as a baseline for comparison with the subsequent models. It is important to note that the model used is the one presented in section \ref{sec3.2}, with 512 neurons on its penultimate dense layer.

\begin{table} 
    \centering
    \begin{tabular}{lrr}
        \toprule
        C3DFC w/ Early Stopping & Accuracy & ROC AUC \\
        \midrule
        AIRTLab  & 95.6\% & 0.9894 \\
        Crowd Violence   & 99.0\% & 0.9994 \\
        Hockey Fights  & 96.6\% & 0.9931\\
        RWF-2000 (400 videos)  & 94.7\% & 0.9922 \\
        \bottomrule
    \end{tabular}
    \caption{Results of the model proposed by \protect\cite{sernani_deep_2021} on different datasets.}
    \label{tab:datasets_results}
\end{table}

\subsection{Model Performance across Datasets and Configurations}

The tables \ref{tab:2epochs} and \ref{tab:1epochs} display the results of training our model on four different databases, with the penultimate layer utilizing 1024 neurons.

\begin{table} [ht!]
    \centering
    \setlength\tabcolsep{5pt}
    \begin{tabular}{lrrr}
        \toprule
        C3DFC w/ One-Cycle & Training & ACC & ROC AUC\\
        \midrule
        AIRTLab   & 3m11s  & 91.0\% & 0.9972 \\
        Crowd Violence   & 59s & 98.4\% & 0.9994 \\
        Hockey Fights   & 1m35s & 94.8\% & 0.9828\\
        RWF-2000 (400 videos)  & 3m03s & 94.9\% & 0.9875 \\
        \bottomrule
    \end{tabular}
    \caption{Results of our model trained for \underline{2 epochs} on four different datasets using super-convergence and a maximum learning rate of $50e^{-2}$. The table shows the training time in minutes, the accuracy (ACC), and the receiver operating characteristic area under the curve (ROC AUC) achieved on each dataset. The model has 1024 neurons in the penultimate layer.}
    \label{tab:2epochs}
\end{table}

\begin{table} [ht!]
    \centering
    \setlength\tabcolsep{5pt}
    \begin{tabular}{lrrr}
        \toprule
        C3DFC w/ One-Cycle & Training & Acc & ROC AUC\\
        \midrule
        AIRTLab   & 1m45  & 89.8\% & 0.9437 \\
        Crowd Violence   & 32s & 97.2\% & 0.9627 \\
        Hockey Fights  & 59s & 94.8\% & 0.9755\\
        RWF-2000 (400 videos)  & 1m52s & 90.4\% & 0.9669 \\
        \bottomrule
    \end{tabular}
    \caption{Results of our model using super-convergence and \underline{1 epoch} of training and a maximum learning rate of $60e^{-2}$.}
    \label{tab:1epochs}
\end{table}

The table \ref{tab:accmulti} shows us the accuracy of the multi-channel models we have experimented with during this research on the complete RWF-2000 dataset \cite{cheng_rwf-2000_2020}.

\begin{table} [ht!]
    \centering
    \begin{tabular}{lrr}
        \toprule
        Context & Flow-gated  & Diff-Gated \\
        \midrule
        Accuracy & 87.25\% & 89.75\%  \\
        Training time & 5h30  & 5h00  \\
        Processing time for a video & 9s & 0.065s\\
        \bottomrule
    \end{tabular}
    \caption{Accuracy, training time, and processing time of our multi-channel input models on the RWF-2000 dataset.}
    \label{tab:accmulti}\label{table:gated_results}
\end{table}

\begin{table}[ht!]
    \centering
    \setlength\tabcolsep{4pt}
    \begin{tabular}{lrrrrr}
        \toprule
        Round & 0 & 1 & 2 & 3 & 4 \\\\
        \midrule
        Accuracy & 50.40\% & 94.84\% &  97.72\%  & 98.80\% & 99.60\%\\
        \bottomrule
    \end{tabular}
    \caption{Accuracy table of our federated model for each round}
    \label{tab:accfed}
\end{table}

Our proposed model, referred to as \textbf{Diff-Gated}, achieves higher accuracy while reducing computation in comparison to the original Flow-Gated architecture \cite{cheng_rwf-2000_2020}. Table \ref{tab:accfed} presents the validation accuracy of our federated model for each round of training on 400 videos from the RWF-2000 dataset. Round 0 accuracy corresponds to the validation accuracy of the model before the first round of training.

\section{Conclusion}
In this paper, we present an innovative examination of machine learning models for violence detection in videos, with a particular emphasis on Federated Learning and our proposed Diff-Gated architecture. We not only maintain or improve upon the accuracy of conventional methods but also reduce training times, enhancing model usability in practical applications.

Our work incorporated super-convergence for transfer learning and explored diverse classifiers for extracting spatio-temporal features from videos. Through modifications to the Flow-gated architecture proposed by \cite{cheng_rwf-2000_2020}, we were able to boost accuracy and cut down training and preprocessing time.

Moreover, we introduced a method for adapting centralized datasets to a federated learning context, using it to train our violence detection model. Despite demonstrating that deep learning models can be effectively trained using federated learning, we note the resource-intensive nature of federated learning, particularly with video data.

While our research has made significant strides, the challenges inherent in federated learning, such as dealing with Non-IID and unevenly distributed data, cannot be overlooked \cite{li2020federated}. Our future efforts will aim at exploring different federated learning strategies and studying the impact of unbalanced client data.

Despite these challenges, our work provides valuable insights for researchers and practitioners alike, underlining the potential of Federated Learning and other novel techniques in violence detection and other real-world applications.

\bibliographystyle{named}
\bibliography{ijcai23}

\end{document}